\documentclass[journal ]{new-aiaa}
\usepackage{graphicx}              % For including images
\usepackage{subcaption}            % For subfigures
\usepackage{caption}               % For customizing captions
\usepackage{listings}
\usepackage[linesnumbered,ruled,vlined]{algorithm2e}

\usepackage[utf8]{inputenc}
\usepackage{textcomp}

\usepackage{graphicx}
\usepackage{amsmath}
\usepackage[version=4]{mhchem}
\usepackage{siunitx}
\usepackage{longtable,tabularx}
\usepackage{xcolor}

\title{Data-driven Probabilistic Trajectory Learning with High Temporal Resolution in Terminal Airspace}

\author{Jun Xiang \footnote{Ph.D. Student, Department of Aerospace Engineering, 5500 Campanile Dr, San Diego, CA 92182, \texttt{jxiang9143@sdsu.edu }, AIAA Student Member} and Jun Chen \footnote{Assistant Professor, Department of Aerospace Engineering, 5500 Campanile Dr, San Diego, CA 92182, \texttt{Jun.Chen@sdsu.edu}, AIAA Senior Member (Corresponding Author)}}
\affil{San Diego State University, San Diego, CA 92182}

\begin{document}
\maketitle
% Abstract
\begin{abstract}

{Predicting flight trajectories is a research area that holds significant merit.
In this paper, we propose a data-driven learning framework, that leverages the predictive and feature extraction capabilities of the mixture models and seq2seq-based neural networks while addressing prevalent challenges caused by error propagation and dimensionality reduction. After training with this framework, the learned model can improve long-step prediction accuracy significantly given the past trajectories and the context information. The accuracy and effectiveness of the approach are evaluated by comparing the predicted trajectories with the ground truth. The results indicate that the proposed method has outperformed the state-of-the-art predicting methods on a terminal airspace flight trajectory dataset. The trajectories generated by the proposed method have a higher temporal resolution(1 timestep per second vs 0.1 timestep per second) and are closer to the ground truth.}
\end{abstract}

\section*{Nomenclature}
% \noindent(Nomenclature entries should have the units identified)

{\renewcommand\arraystretch{1.0}
\noindent\begin{longtable}{@{}l @{\quad=\quad} l@{}}
$b$ & branching factor of the graph of the A* task\\
$C_a$ & arrival cost \\
$comt_j$ & computing time of the jth cycle\\
$d$ & solution depth  of the A* task\\
$d_h$  & a parameter set by the user used to decide the search algorithm indicator \\
$d_t$ & destination of agents\\
$dt_j$ &actual processing time for the jth cycle\\
$d\alpha$  & increment of $\alpha$ \\
$d\beta$  & increment of $\beta$ \\
$e$ & error of the A* tas\\
$\mathbf{E}$   & edge set connects the node \\
$Ed$   & Euclidean distance \\
$e_{msa*}$  &  edge cost \\
$G$  & grid graph \\
$g_{xy}^s$ &    unit grid cell, located at x in the x direction, y in the y direction, with status s\\
$\mathbf{N}$ & full quadtree node set\\
$n_{l,o}$ & the node is in the full quadtree node set at the level l and its center position is o \\
$p$  & current location of the agent \\
$R$& risk value function \\
$s$ & search algorithm indicator\\
$t_h$ & a parameter set by the user used to decide the search algorithm indicator \\
$T_k$ & flight plan for agent k \\
$x_r$ & a parameter set by the user controls how often the $\alpha$ and $\beta$ are decremented\\
$\alpha$  & a parameter set by the user used to determine a node should be divided \\
$\alpha_h$ & a parameter set by the user used to decide the search algorithm indicator \\
$\beta$  & a parameter set by the user used to determine a node should be divided \\
$\tau$ & time step \\
$\omega$   & arrival cost coefficient \\

% \multicolumn{2}{@{}l}{Subscripts}\\
% cg & center of gravity\\
% $G$ & generator body\\
% iso	& waypoint index
\end{longtable}}

\section{Introduction}\label{sec: intro}
The terminal airspace is the airspace around airports. This airspace is characterized by low altitudes, multiple agents interacting closely, dynamic conditions, and the need for fast decisions. However, only 4\% of all active airports in the US have control towers that act as a centralized authority for separation\cite{patrikar2022predicting}. In these non-towered airspaces, pilots bear the sole responsibility for ensuring their aircraft's safety, and the risk of collision is higher. Therefore, predicting the trajectories of other flights in terminal airspace becomes more important.

Furthermore, it is more crucial to track and predict the motion of UAVs, since more and more Unmanned Aerial Vehicles (UAVs) are occupying the airspace when the technology for autonomy has evolved and matured~\cite{su2021study}.
The Federal Aviation Administration (FAA) projects a drone fleet size of 1.81 million units by 2026~\cite{FAA}.  The security service needs to anticipate the misuse of UAVs in public areas, which includes public demonstrations, mass events, and airports~\cite{laurenzis2020prediction}.
Most of the UAVs should be under surveillance by the Unmanned Aircraft System Traffic Management (UTM), which are denoted as in-network UAVs. However, those in-network UAVs could turn into out-network UAVs occasionally due to the limitations and loss of communication~\cite{zhang2023communication}. Therefore, timely and accurate trajectory prediction is the key to preventing potential collisions with out-network UAVs. This is even more critical for conflict detection and resolution in UAV landing management~\cite{wu2022safety}, which often occurs near crowded airspace~\cite{zhou2020optimized}.

% Concurrently, UAVs under Air Traffic controller need to predict the trajectory of out-network UAVs to prevent collision, which is especially important in UAV landing management owing to the crowded airspace near vertiports~\cite{zhou2020optimized}. 
% Even though most of the UAVs are in-network, the communication capabilities are also limited~\cite{zhang2023communication}. Therefore, in-network UAVs may suddenly turn into out-network UAVs. Trajectory prediction of in-network UVAs is also critical while conflict detection and resolution technology, which are significant functions in the air traffic system~\cite{lin2018uav,wu2022safety}, highly relies on accurate and timely predictions of out-network UAV trajectories~\cite{kang2020model}. 

Machine learning (ML) and data-driven methods have demonstrated impressive predictive capabilities across various domains, transcending traditional areas such as motion planning~\cite{zhang2024self}, cybersecurity~\cite{jin2024online, weng2024leveraging, wu2024application}, and health~\cite{ni2024timeseries,yukun2019deep,Li2024.09.08.24313212}. Furthermore, Generative Pretrained transformer (GPT) achieves promising results on multiple NLP tasks such as question answering, automatic summarization, and sentiment analysis~\cite{brown2020language}
Contrastive Language–Image Pre-training (CLIP) dramatically boosts self-supervised learning and achieves promising results on multiple CV tasks such as image classification~\cite{radford2021learning}. The Vision Transformer (ViT)~\cite{dosovitskiy2020image}, and its extended work~\cite{xin2024vmt} proposed a transformer-based approach for handling computer vision tasks by treating images as sequences of fixed-size patches. By learning patterns and relationships from large amounts of historical data, ML algorithms can generate accurate predictions in real time. These algorithms may be enhanced by incorporating additional information including weather patterns, air traffic patterns, and other contextual factors that may impact UAV motion~\cite{chen2023bridging}. TSM~\cite{ni2024timeseries} demonstrate the superior performance of transformer-based models, particularly PatchTST, in capturing the complex patterns in time series data, and their insights have been crucial in guiding later approaches to time series signal analysis. 
3D trajectory data have many similar features with images and text. The raw trajectory data have the same dimension as the raw image data. Both the trajectory data and text are temporal data. Therefore, data-driven and ML are suitable to be applied in the trajectory prediction of many objects including aircraft.

In this paper, a learning framework is proposed to provide high-temporal-resolution trajectory prediction while maintaining high accuracy. This paper's primary contributions are as follows: 
\begin{itemize}
\item  The proposed learning framework can improve long-step prediction accuracy significantly on a terminal airspace dataset, outperforming contemporary state-of-the-art methods.
\item  Our work leverages the predictive and feature extraction capabilities of the mixture models and seq2seq-based Neural Networks while addressing prevalent challenges caused by error propagation and dimensionality reduction.
\item It is the first framework to propose mixture models learn from the output of neural networks to predict flight trajectories. Many researchers have applied neural networks and mixture models together to predict trajectories, however, all the previous methods simply generate the mixture model parameters, including weights, means, and covariances for each mixture component with neural network~\cite{huang2019uncertainty}~\cite{shi2022motion}.

\end{itemize}

\section{Related Work}
Trajectory prediction for aircraft is a trending research topic with many proposed methods. Most predictors leverage model-based, cluster-based, and ML-based methods. The most traditional way is predicting future aircraft motion naively by physical equations of motion~\cite{chatterji1996route}. 
The 3D hybrid model in~\cite{wang20193} combines both data-driven and physics models, so it is capable of predicting UAV motion using minimal real flight data.
Recently, the hidden Markov model (HMM) has been utilized to forecast trajectories, taking into account environmental uncertainties~\cite{ayhan2016aircraft}.
By using the Bayesian network, the statistical patterns of aircraft dynamics can be comprehended to predict their further motion~\cite{kochenderfer2010airspace}.
%Through the Bayesian network, one can comprehend the statistical patterns of dynamic aircraft variables and predict their motion~\cite{kochenderfer2010airspace}. 
The work in~\cite{barratt2018learning} uses the probabilistic generative model to predict the flight trajectory near the terminal by directly learning from aircraft positional data.
% The probabilistic generative model predicts the aircraft trajectory near the terminal by learning from positional data~\cite{barratt2018learning}. 
TrajAir~\cite{patrikar2022predicting} fuses the weather and socially past trajectories data with attention neural network and generates prediction. Recurrent Neural Network (RNN) can be trained on historical UAV data to learn the underlying patterns in the motion of UAVs. The trained RNN can then be used to predict future motion based on current position and velocity information. A proposed RNN is designed to forecast aircraft trajectories incorporating weather characteristics~\cite{pang2019recurrent}, which are well suited for time series prediction problems. Moreover, a 4D aircraft trajectory prediction model is proposed in~\cite{liu2018predicting}, which leverages the deep generative convolutional RNN framework.
% The deep generative convolutional RNN approach is proposed for aircraft trajectory prediction in 4D~\cite{liu2018predicting}. 
Generative Adversarial Networks (GAN), which include a discriminator and a generator, can also generate flight trajectory~\cite{xiang2023landing}.  A Multilayer Perceptron Neural Network machine learning model predicts the deviations in actual flight paths from the planned routes, by extracting features from historical surveillance data that can be generalized for different route structures~\cite{zhu2023predicting}. A Conditional tabular generative adversarial network(CTGAN)\cite{zhang2024four} model is trained and tested using historical 4D trajectory data from the Hong Kong region. This study\cite{schimpf2023generalized} compared hybrid-recurrent neural network models(CNN-LSTM, CNN-GRU, and SA-LSTM) on aircraft trajectory prediction. Pang et al.\cite{pang2021data} used an advanced Bayesian deep learning approach for predicting aircraft trajectories, incorporating the effects of weather.

Besides aircraft trajectory prediction, many data-driven and ML-based methods are also proposed to predict the trajectory of other objects. In many previous trajectories predicting methods, the Gaussian Mixture Model (GMM) is a popular method to represent the distribution of predicted trajectories~\cite{shi2022motion}\cite{fabisch2021gmr}\cite{varadarajan2022multipath++}, because of its capability to model multimodal behaviors of the predicting agents.
%In many previous trajectories predicting methods, because the behaviors of the agents are highly multimodal, the Gaussian Mixture Model(GMM) is commonly used to represent the distribution of predicted trajectories~\cite{shi2022motion}\cite{fabisch2021gmr}\cite{varadarajan2022multipath++}. 
A proposed extended Kalman filter is used to estimate vessel states which is subsequently used to help predict the vessel trajectories~\cite{perera2012maritime}. The representation of interactions through graphs is also garnering increasing interest, which results in the application of graph neural networks (GNN) for trajectory prediction~\cite{mo2022multi}.
An HMM trajectory prediction algorithm is also developed with a self-adaptive parameter that can predict trajectory in a network-constraint environment~\cite{qiao2014self}. 

In conclusion, many techniques have been proposed for predicting flight trajectories. Nevertheless, current methods for predicting flight trajectories are typically very slow and not suited for online tasks~\cite{xie2022efficient}. Meanwhile, all the previous trajectory prediction methods still do not meet the accuracy requirement for many applications. Besides, many planning methods struggle to predict the uncertainty of future trajectories. A flight may have multiple different path options, but some predictors can not anticipate them. 
% \textcolor{red}{I think you need a short sentence to summarize the shortcomings of those methods.}

\section{Problem Statement}\label{Sec: ps}
The goal of this paper is to predict the trajectory of flights in the near future using their past trajectories, and any useful given information such as wind information and their neighbors. In this section, the inputs and outputs are going to be defined.

The flight trajectory usually is discretized into multiple waypoints. We locate the physical flight position in airspace with three-dimensional Cartesian coordinates. Considering time, each point is on a 4-D space defined by 4 variables: time(t), the position in the x-axis (x), the y-axis (y), and the z-axis (z) of the airspace. Usually, z is the altitude of the drone, and (x, y) is the position in the plane that is parallel to the ground. Therefore, the past trajectory of flight f, $T_P^f$, is expressed as:
\begin{equation}
    T_P^f = \{p_1^f, p_2^f, p_3^f....p_n^f\} \in \mathbb{R}_{n\times4} 
\end{equation}
\begin{equation}
    p_i^f = <t_i, x_i^f, y_i^f, z_i^f> \in \mathbb{R}_{4} 
\end{equation}
where $f\in\mathcal{F}=\{1,2,...f,...F\}$ is one of the flight in the current airspace. $T_P^f$ contains points starting from time $t_1$ to time $t_n$. The aircraft flies from the position $(x_1, y_1, z_1)$ to the position $(x_n, y_n, z_n)$ in the air space within the time interval $t_1$ to $t_n$.
The future trajectory($T_F^f$) has the same form as the past trajectory. 
\begin{equation}
    T_F^f = \{p_{n+1}^f, p_{n+2}^f, p_{n+3}^f...p_{n+k}^f\} \in \mathbb{R}_{k\times4} 
\end{equation}
The time set in the future trajectory, $t_F = \{t_{n+1}, t_{n + 2}...t_{n+k}\}$, is decided by the specific task based on which specific time point of the position is desired to be predicted.
\begin{figure}[!h]
    \centering
    \includegraphics[width=0.48\textwidth]{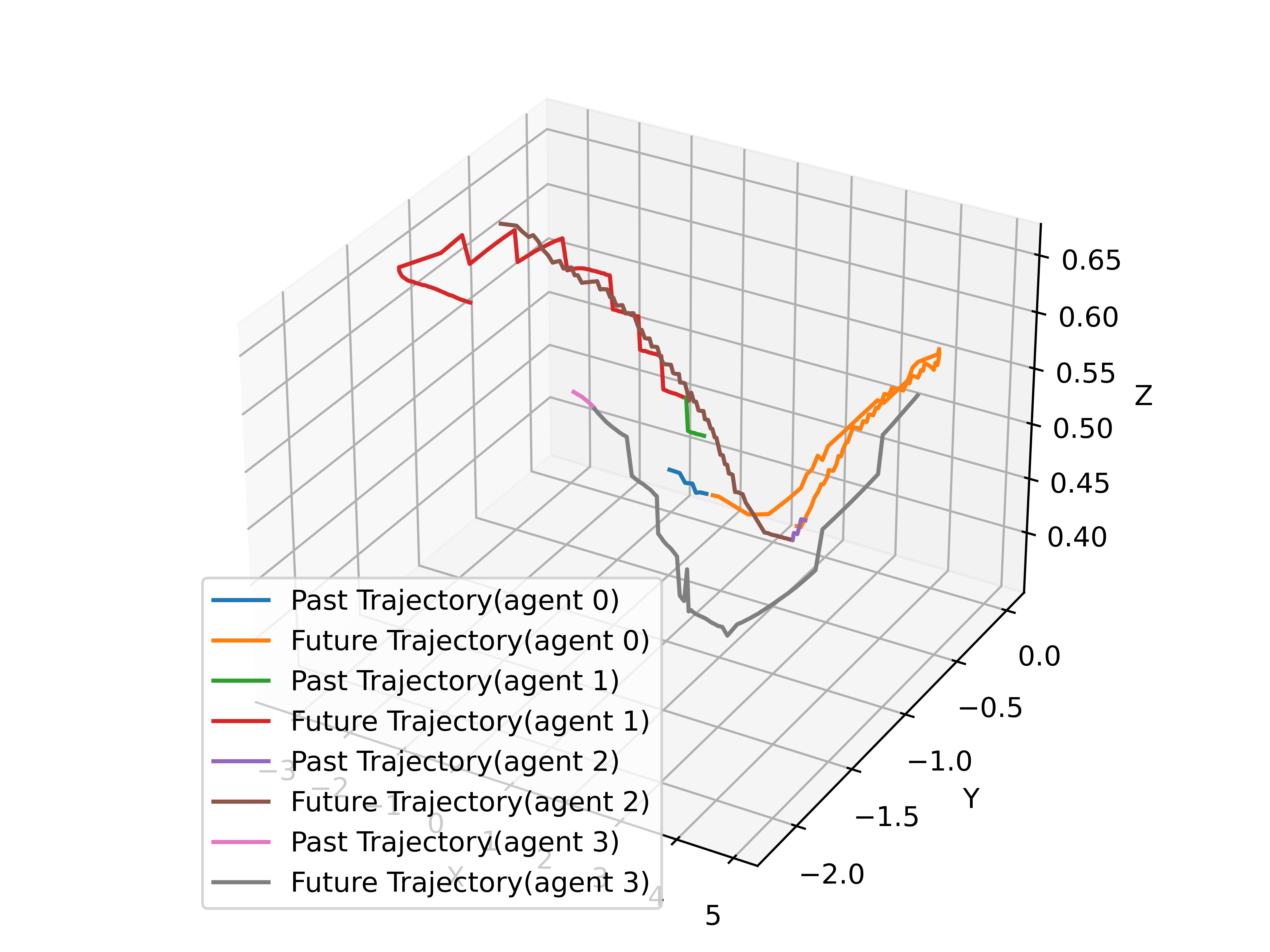}
    \caption{Past trajectories and future trajectories}
    \label{fig:realT}
\end{figure}
Figure~\ref{fig:realT} shows an example of four agents and their past and future trajectories $T_p$ and $T_f$ in the airspace.

The predictor $G$ we proposed should take the past trajectory of the ego agent $T_P^e$, the context information nearby $\phi^e$, and the past trajectory of the flights near the agent $T_P^f, \quad \forall f\in\{1, \ldots, F\}$ as input, and output the ego agent's the predicted trajectory $T_{pred}^e$. Here, the ego agent is the agent of interest for which we want to predict its future trajectory.

\begin{equation}
T_{pred}^e  = G(T_P^e, T_P^f, \phi^e)
\end{equation}
%For a trajectory predicting problem, achieving 100\% accuracy is very unlikely. No previous work has ever achieved that. Therefore, Instead of predicting the points of trajectory directly, a probability distribution of the points will be predicted.

%G denotes the trajectory prediction generator, the trajectory prediction generator takes the past trajectory as input, and generates $\mathcal{P}_i$, the probability distribution of the flight f position at time i. 
The goal of training the predictor $G$ is to minimize the prediction error $\Delta$ between the predicted trajectory and the real future trajectory as shown in Eqn~\ref{eq: error}. The detailed performance metric for prediction error in application will be defined in the result section.
\begin{equation}
\underset{\theta}{\operatorname{Min}} \quad \Delta(T_{pred}^e, T_F^e)
\label{eq: error}
\end{equation}
where $\theta$ is the learning parameter.
Moreover, because the prediction result will be further used for the subsequent tasks, such as collision avoidance, air traffic management, and path optimization, and the flight flies extremely fast, the prediction process should be faster enough so that more time can be left for the subsequent tasks. 
\begin{equation}
    O(G) << t_{n + 1} - t_{n}
\end{equation}
where O(G) is the computing time of the prediction process, $t_n$ is the time we start to predict, and $t_{n+1}$ is the time when the desired forecast trajectory starts.

In other words, the trajectory predictor should be able to take the past trajectory of the ego agent and all the related context features and predict the future trajectory in a very short time.

\section{High Temporal Resolution Prediction Framework}

\subsection{Challenges for high-temporal-resolution prediction framework}

\begin{figure}
    \centering
    \includegraphics[width=0.48\textwidth]{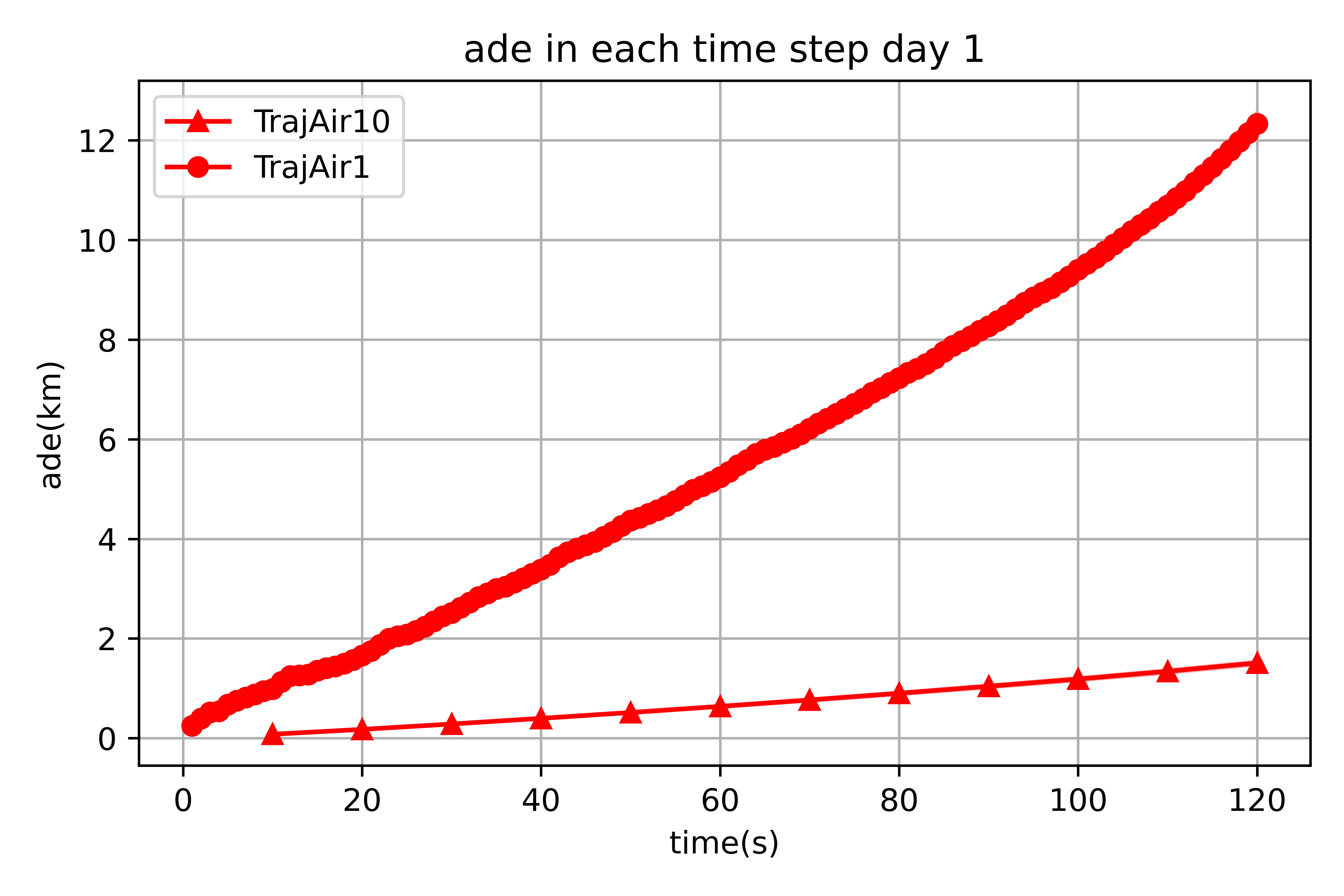}
    \caption{Feedforward neural networks suffer from error propagation(Average Displacement Error (ADE) increases very fast along the time)}
    \label{fig:FN}
\end{figure}

\begin{figure*}
    \centering
    \includegraphics[width=0.9\textwidth]{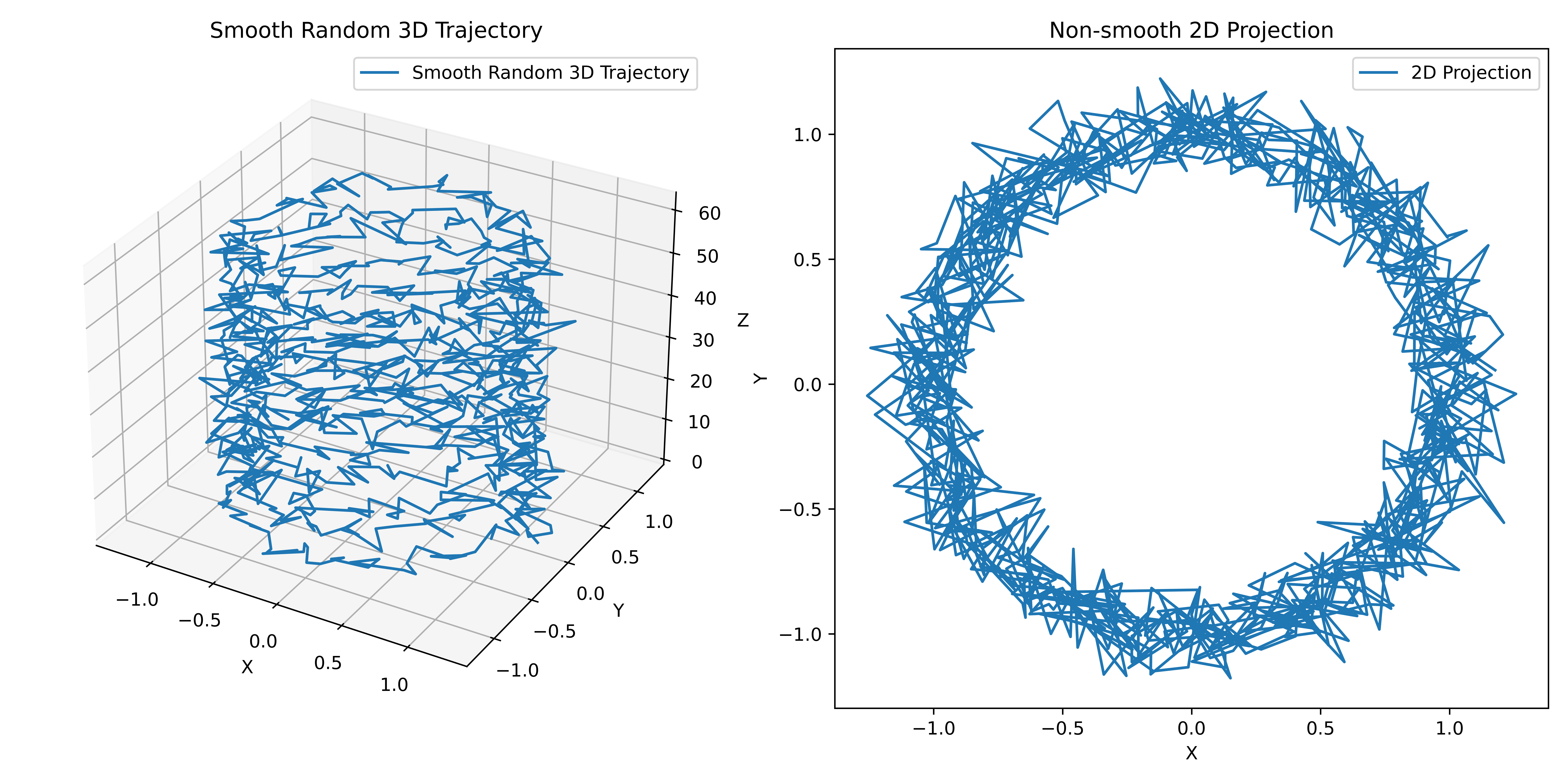}
    \caption{Smooth 3D trajectory can have an unsmooth 2d projection}
    \label{fig:smooth}
\end{figure*}

\begin{figure}
    \centering
    \includegraphics[width=0.48\textwidth]{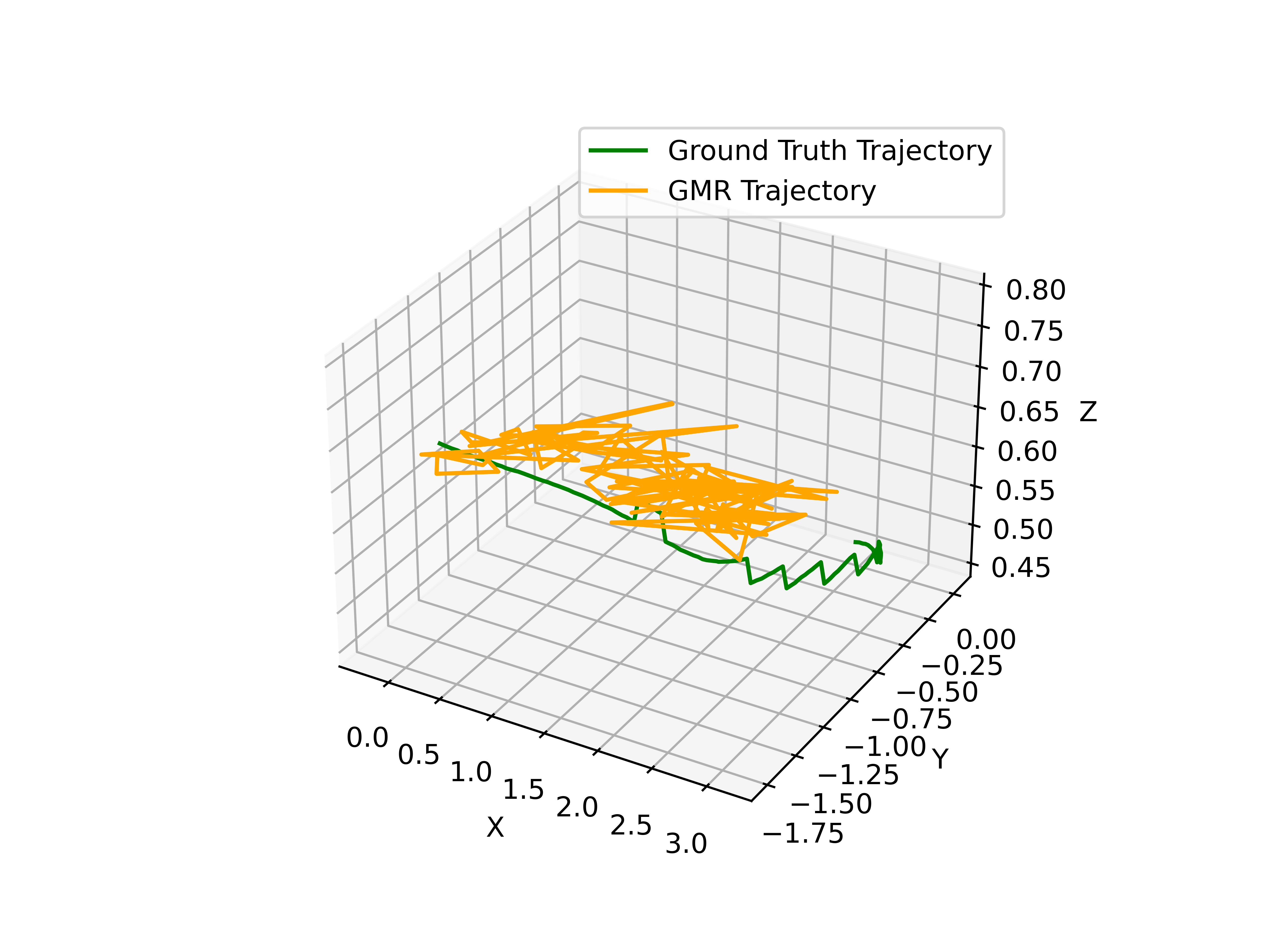}
    \caption{High-dimension mixture model regression produces an unsmooth 3-D trajectory}
    \label{fig:gmr4}
\end{figure}

High-density autonomy operations require a high-temporal-resolution trajectory prediction model. In statistical words, a mixture model signifies the mixture distribution that represents the probability distribution of observations across the entire population. The mixture model is one of the most promising solutions, because the future trajectory of the flights is highly heterogeneous, a mixture of probabilistic models can be very efficient in representing the future trajectory. Mixture models have been adopted in many trajectory-predicting projects and achieved significant results~\cite{barratt2018learning}~\cite{shi2022mtr}. Compared to the output of deep learning models, the mixture model does not suffer from the same kind of error propagation issues that are often associated with feedforward neural networks or other deep learning models. Figure~\ref{fig:FN} shows an example of a most recent feedforward neural network model, called TrajAir~\cite{patrikar2022predicting} that has a much larger error when predicting high-temporal-resolution trajectories. The identical neural network-based predictor performs much worse when generating results for every time step (i.e. TrajAir1 in Fig.~\ref{fig:FN}) compared to generating results for every 10-time steps (i.e. TrajAir10 in Fig.~\ref{fig:FN}). On the other hand, the naive mixture model also presents challenges when predicting trajectories because it relies on context information, where the given features are large-scale and high-dimensional. Dealing with high-dimensional features can be challenging due to the peaking phenomenon~\cite{hughes1968mean}, which means the performance of a machine-learning model eventually decreases as the dimensionality of the feature space increases. However, research~\cite{zollanvari2020theoretical} has shown that the predictive power of the mixture model increases with additional features as long as the net effectiveness of an added feature set exceeds the size of this added feature set. Moreover, although mixture models are capable of clustering and generating high-dimension data, we observed that the 3-D trajectories of the flight generated from high-dimension data by the mixture model are not smooth. As shown in Figure~\ref{fig:smooth}, the 2-dimension projection of the trajectory is not smooth even though the same trajectory in the 3-dimension is smooth. Figure~\ref{fig:gmr4}  shows an example of a nonsmooth predicted trajectory generated by high-dimension mixture model regression. Even each point of the generated trajectory is relatively to the ground truth trajectory, the generated trajectory is not useful. All in all, high-dimensional features are crucial for prediction accuracy, but directly incorporating them into a mixture model proves ineffective.
Therefore, in our framework, we proposed the neural network guide-generating module to reduce the dimension of the given information while keeping the useful information as much as possible.
the training dataset for the mixture model then only includes three dimensions(x, y, and z in the Cartesian coordinate), avoiding nonsmoothness caused by projection process. 

\begin{figure*}[t]
    \centering
    \includegraphics[width=0.8\textwidth]{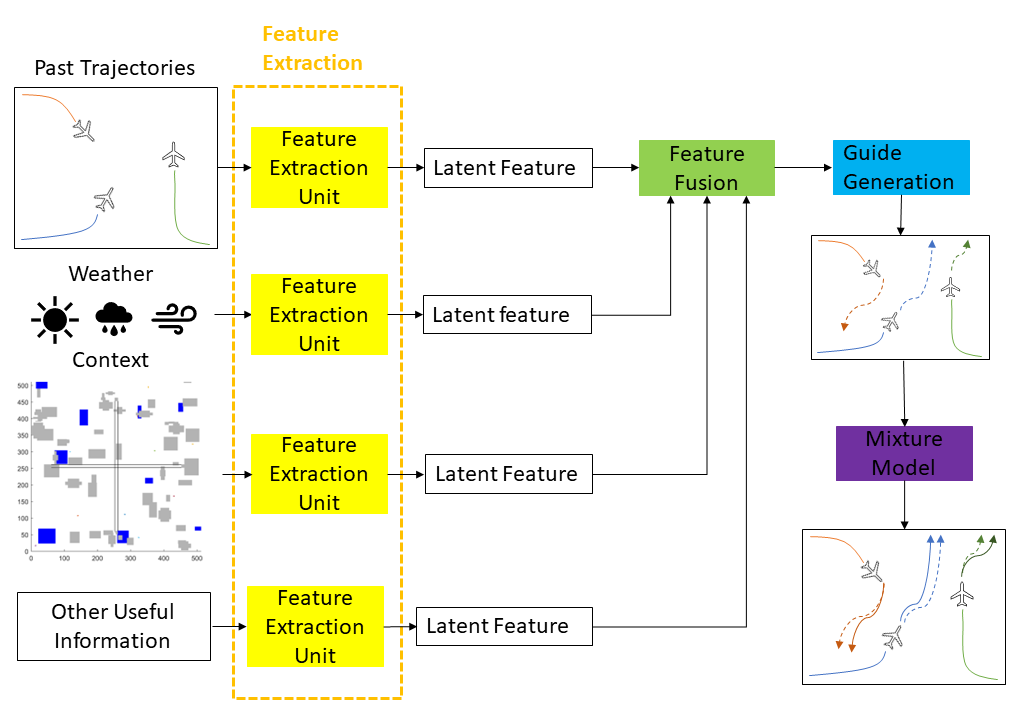}
    \caption{Proposed learning framework}
    \label{fig:framework}
\end{figure*}

Therefore, to overcome the above challenges to provide a high-temporal-resolution prediction framework, we propose a learning framework, as shown in Fig~\ref{fig:framework}, that consists of four main modules: feature extraction, feature fusion, neural networks guide generator, and mixture model predictor. The raw input includes the past trajectory learning models. Information about the ego agent, the past trajectory of other agents, and context information are sent to feature extraction units in the feature extraction modules separately. After the feature extraction units extract important features from the raw input, the feature fusion module combines those features from all the raw input together. The guide generator then generates the trajectory guide using the fused feature, which contains the information from all the inputs. Finally, the mixture model predictor module generates the trajectory prediction based on this guide. 

\begin{figure*}[h]
    \centering
    \includegraphics[width=0.8\textwidth]{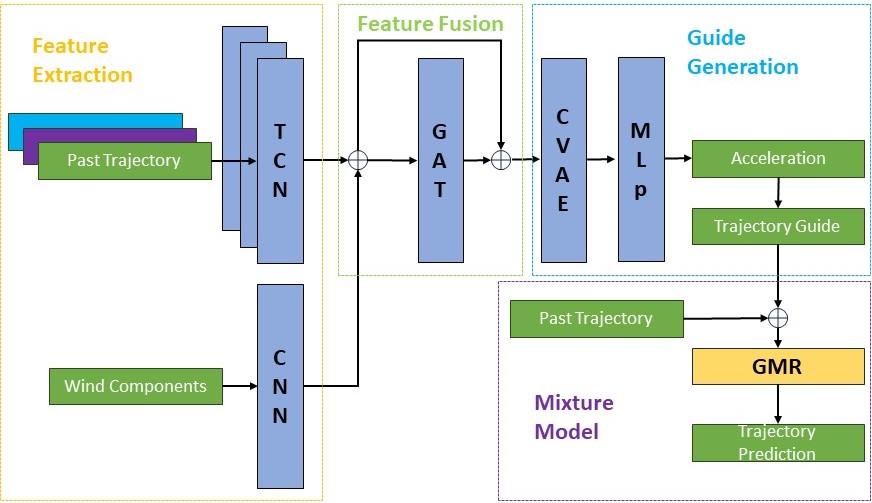}
    \caption{Implementation details of the framework}
    \label{fig:implementation details}
\end{figure*}
In this paper, we trained a future trajectory predictor with our framework on a general aviation trajectory dataset~\cite{patrikar2022predicting}, which uses the past trajectories of the ego agent, the past trajectories of the neighboring agents, and weather content as input to predict the ego agent's future trajectory. We pick neural networks from the TrajAirNet model~\cite{patrikar2022predicting} and probabilistic trajectory models~\cite{barratt2018learning} to complete the framework. 
The details of the implementation of our learning framework are shown in Fig.~\ref{fig:implementation details}. Each module is introduced in detail as follows.

\subsection{Feature extraction module}
{Feature extraction module} transforms all the provided raw data which may affect the future trajectory into a set of features that effectively represent the underlying characteristics of the data and yield improved results compared to applying machine learning directly to the raw data. Feature extraction is a crucial step in the data-driven pipeline. Raw data in different formations needs different extraction methods. In this framework, neural network-based methods are preferable, because the features extracted result are easier to be integrated into the neural network's guide generator module. In the trajectory predicting tasks, the most commonly provided raw data is the past trajectory, which is a form of sequential data. The Long Short-Term Memory (LSTM) networks~\cite{chen2022intention} are useful for extracting features from past trajectories. Another way to extract features from past trajectories is the Autoencoders~\cite{ivanovic2020multimodal}. Temporal Convolutional Networks (TCNs)~\cite{patrikar2022predicting} are also applied to handle the past trajectories. Another common raw input is the picture of the environment near the ego agent. Graph neural networks (GNNs)~\cite{gao2020vectornet} is an option to extract features from the picture for trajectory prediction tasks. 

In this paper, the feature extraction module consists of two input layers which are Temporal Convolutional
Networks (TCNs), and Convolutional Neural Networks (CNN). The past trajectories of both ego and neighbor agents will be sent to TCNs and the wind context will be sent to CNNs. The TCN~\cite{bai2018empirical} layers can convert the spatiotemporal trajectory into a latent vector while ensuring that causal connections within the data are preserved. The TCNs simply consist of two layers of dilated causal convolution and rectified linear unit (ReLU), which provide non-linearity, followed by a spatial dropout.
\begin{equation}
h_{tcn}^f=T C N\left(T^f_P\right), \forall f \in\{1, \ldots, F\}
\end{equation}
The CNN was widely recognized since~\cite{lecun1998gradient}. The CNNs can combine the wind context near the different flights together and transfer the feature to a shape we want. The wind near flight f consists of wind velocity in x and y direction.
\begin{equation}
    w^f = (wvx, wvy)\in \mathbb{R}_{2}
\end{equation}
Then, in the feature fusion module, the output of TCNs and CNNs are concatenated together and become the encoded hidden item.
\begin{equation}
h_{e n c}^f=h_{tcn}^f \oplus C N N\left(w^f\right) \forall f \in\{1, \ldots, F\}
\end{equation}

\subsection{Feature fusion module}
{Feature fusion module} combines data from different sources or forms to create a more informative and selective feature representation. It can also combine features from different layers or branches. It is extremely important to notice that the selection of the feature fusion module should be closely aligned with the feature extraction module. Simple concatenation~\cite{patrikar2022predicting} is a traditional feature fusion that is suitable for the same feature space. Pooling~\cite{gupta2018social} also has been used to fuse interaction information in trajectory prediction. Short skip connections fuse the identity mapping features and residual learning features in In Residual Networks (ResNet)~\cite{he2016deep}~\cite{chen2021multimodal}.
Inception, introduced in~\cite{szegedy2015going}, fuses the outputs of a variety of convolutions and pooling layers. Some research fuses outputs of multiple branches with different kernel sizes with Attention~\cite{dai2021attentional}~\cite{li2019selective}. At this point, convolutional mechanism~\cite{hu2020aesthetic} and attention mechanisms are stand-out neural network structures for feature fuse in trajectory prediction tasks. Many previous papers have applied convolutional mechanisms~\cite{song2020pedestrian} and attention mechanisms~\cite{zhang2022ai} to fuse and extract common information in trajectory prediction tasks.  

In this paper, we choose graph attention network (GAT)~\cite{velivckovic2017graph} as the main neural network in the feature fusion module. The GAT layer can decide which information is more important than others in the encoded hidden item which extract feature from the past trajectory and wind context of all flights. In a scalable way, an attention mechanism, as seen in GATs, can selectively focus on specific agents that affect the trajectory of the ego agent. 
\begin{equation}
h_{g a t}^{f}=G A T\left(h_{e n c}^f\right)
\end{equation}
The GATs generate a hidden output for each agent. The hidden output is then concatenated and segregated with the previously encoded hidden item of the ego flight $h_{e n c}^e$.

\subsection{Neural networks guide generator}
{Neural networks guided generator} can generate the guide for the mixture probabilistic module giving fused features from the feature fusion module. After the raw data is extracted and fused together, the processed data will be sent to the guide-generating module. The guide-generating module can be an output layer in the Neural Networks. Some research~\cite{zhao2021tnt} simply uses MLP as the output layer to generate the trajectories. Another popular generating output layer is sequential generating such as the output layer of LSMT~\cite{alahi2016social}, RNN~\cite{pang2019recurrent}, and Transformer~\cite{giuliari2021transformer}.  Variational Autoencoders (VAEs)~\cite{kingma2019introduction} is another type of generative model that generates new data samples that are similar to the ones in the training dataset.

In this paper, we construct a Conditional Variational Autoencoders (CAVE)-based neural network as the guide generator~\cite{kingma2014semi}. The concatenated vector from the feature fusion module is sent to CAVE, which is the first layer of the neural networks guide generator, as the conditional input. 
The CVAE consists of two main parts, an encoder Q(·) and a decoder P(·). After training, the decoder can take the concatenated vector and a Gaussian noise $z$ and generate a probability distribution. The generated probability distribution then can sample a hidden output for the ego agent $h_{cvae}^e$. 
\begin{equation}
\begin{aligned}
z & \sim \mathcal{N}(0, I) \\
h_{\text {cvae }}^e & \sim P\left(h_{\text {cvae }}^e \mid z, h_{\text {enc }}^e \oplus h_{\text {gat }}^f\right)
\end{aligned}
\end{equation}
The output layer is MLP, the MLP take the sampled CVAE output $h_{cvae}^e$ and output the acceleration output. The dimension of the output will be (k/$\Delta$ t, 1). $\Delta$ t is the time interval between each time step. k is the task prediction length. Each element in the output vector represents the acceleration $s^e$ of the ego agent in each time step from $t_{n + 1}$ to $t_{n + k/\Delta t}$.
\begin{equation}
s_{t_{n + 1}: t_{n + k/\Delta t}}^e=M L P\left(h_{\text {cvae }}^e\right)
\end{equation}
The acceleration output is subsequently transformed into absolute positions simply with the kinematic equation. And the calculated absolution positions can form the trajectory guide $T_{guide}^{e}$. 
\begin{equation}
p_{t+1}^{e}=2 p_t^{e}-p_{t-1}^{e}+p_t^{e} \Delta t^2
\end{equation}
\begin{equation}
T_{guide}^{e}=\{p_{n+\Delta t}^{e}, p_{n+2\Delta t}^e, \ldots, p_{n+(k/\Delta t)\Delta t}^{e}\}
\end{equation}
We call the $T_{guide}^{e}$ trajectory guide because it is then used to guide the mixture model predictor module to generate the prediction trajectory.

\subsection{Mixture model predictor module}

The mixture model predictor module is the core module of this framework that will generate the final trajectory prediction given the guide generated by the guide-generating module.
In this paper, Gaussian Mixture Regression(GMR) is deployed as the mixture model predictor module.
The GMR predictor is initially introduced in~\cite{fabisch2021gmr} and ~\cite{calinon2007learning}. The specific one employed in this paper is proposed by~\cite{barratt2018learning}. It is one of the most widely recognized trajectory predictors. Before predicting, we need to train a Gaussian Mixture Model (GMM). First, the GMM learns the training dataset through the Expectation Maximization (EM)~\cite{dempster1977maximum} algorithm. A trajectory of the training dataset consists of three parts, the first part is the past trajectory $T_p^e$, the second part is the future trajectory guide generated by the previous neural networks guide generator $T_{guide}^e$, and the ground truth future trajectory $\hat{T}_{F}^e$. After the GMM is trained, the GMM is shown as equation~(\ref{gmrt}).

\begin{equation}\label{gmrt}
p(\boldsymbol{x}, \boldsymbol{y})=\sum_{k=1}^K \pi_k \mathcal{N}_k\left(\boldsymbol{x}, \boldsymbol{y} \mid \boldsymbol{\mu}_{\boldsymbol{x} \boldsymbol{y}_k}, \boldsymbol{\Sigma}_{\boldsymbol{x} \boldsymbol{y}_k}\right)
\end{equation}
where $\mathcal{N}_k\left(\boldsymbol{x}, \boldsymbol{y} \mid \boldsymbol{\mu}_{\boldsymbol{x} \boldsymbol{y}_k}, \boldsymbol{\Sigma}_{\boldsymbol{x} \boldsymbol{y}_k}\right)$ are Gaussian distributions with mean $\boldsymbol{\mu}_{\boldsymbol{x} \boldsymbol{y}_k}$ and covariance $\boldsymbol{\Sigma}_{\boldsymbol{x}\boldsymbol{y}_k}$, $K$ is the number of Gaussians, and, x is the input and y is the output, where x is the combined past trajectory and future trajectory guide of the ego agent $(T_p^e, T_{guide}^e) \in R_{(n + k/\Delta t) \times 4}$. y is the trajectory prediction $T_{pred}^e \in R_{k\times4}$. $\pi_k \in [0, 1]$ are priors that sum up to one. If $\pi_k$ is larger, the more likely the (x, y) belongs to $\mathcal{N}_k$. 

GMR estimates the distributions of the output $\boldsymbol{y}$ with a given input x by determining the conditional distribution $p(\boldsymbol{y} \mid \boldsymbol{x})$,  The conditional distribution for each individual Gaussian is defined by
\begin{equation}
\mathcal{N}\left(\boldsymbol{x}, \boldsymbol{y} \mid  \boldsymbol{\mu}_{x y}, \boldsymbol{\Sigma}_{x y}\right) 
\end{equation}

where

\begin{equation}
\boldsymbol{\mu}_{\boldsymbol{x} y}=\left(\begin{array}{c}
\mu_x \\
\mu_y
\end{array}\right), \quad \boldsymbol{\Sigma}_{x y}=\left(\begin{array}{cc}
\boldsymbol{\Sigma}_{x \boldsymbol{x}} & \boldsymbol{\Sigma}_{\boldsymbol{x} y} \\
\boldsymbol{\Sigma}_{y x} & \boldsymbol{\Sigma}_{y y}
\end{array}\right)
\end{equation}

\begin{equation}
\mu_{y \mid x}=\mu_y+\Sigma_{\boldsymbol{y} x} \Sigma_{x \boldsymbol{x}}^{-1}\left(x-\mu_x\right) \\
\end{equation}
\begin{equation}
\Sigma_{y \mid x}=\Sigma_{y y}-\Sigma_{y x} \Sigma_{x \boldsymbol{x}}^{-1} \Sigma_{x y}
\end{equation}

 the prior and conditional distribution of each individual Gaussian is determined by the following equation
\begin{equation}
\pi_{\boldsymbol{y} \mid \boldsymbol{x}_k}=\frac{\mathcal{N}_k\left(\boldsymbol{x} \mid \boldsymbol{\mu}_{\boldsymbol{x} k}, \boldsymbol{\Sigma}_{\boldsymbol{x} k}\right)}{\sum_{l=1}^K \mathcal{N}_l\left(\boldsymbol{x} \mid \boldsymbol{\mu}_{\boldsymbol{x} l}, \boldsymbol{\Sigma}_{\boldsymbol{x} l}\right)}
\end{equation}

after the priors are calculated, we can obtain the conditional distribution with the following equation
\begin{equation}
p(\boldsymbol{y} \mid \boldsymbol{x})=\sum_{k=1}^K \pi_{\boldsymbol{y} \mid \boldsymbol{x}_k} \mathcal{N}_k\left(\boldsymbol{y} \mid \boldsymbol{\mu}_{\boldsymbol{y} \mid \boldsymbol{x}_k}, \boldsymbol{\Sigma}_{\boldsymbol{y} \mid \boldsymbol{x}_k}\right)
\end{equation}
once we have the conditional distribution, we can simply sample an output y from the conditional distribution as the final prediction $T_{pred}^e$.

\subsection{Loss function and accuracy of trajectory prediction}
The entire neural networks pipeline uses integration loss functions
\begin{equation}
\mathcal{L}_{\text {total }}=\mathcal{L}_{\text {traj }}+\mathcal{L}_{\text {cvae }}
\end{equation}
The $\mathcal{L}_{\text {traj }}$ measures the mean squared error between the generated guide future trajectory and the ground truth.
\begin{equation}
\mathcal{L}_{t r a j}=\operatorname{MSE}\left(T_{guide}^{e}, \hat{T}_{guide}^{e}\right)
\end{equation}
The $\mathcal{L}_{\text {cvae }}$ quantifies the KL-Divergence between the sampling distribution of the latent variable,  $\mathcal{N}(0, I)$, and the distribution of the trained latent variable.
\begin{equation}
\mathcal{L}_{\text {cvae }}=D_{k l}\left(Q\left(z \mid \hat{h}_{guide}^e, h_{\text {enc }}^e \oplus h_{\text {gat }}^e\right) \| \mathcal{N}(0, I)\right)
\end{equation}
The hidden item $\hat{h}_{guide}^e$ is the hidden output of TCNs take actual ground truth trajectory guide of the ego agent. If the output of the Neural Networks is closer to ground truth, the KL-Divergence should be smaller.
\begin{equation}
\begin{aligned}
\hat{h}_{guide}^e & =T C N(\hat{T}_{guide}^{e} ) \\
\end{aligned}
\end{equation}

To measure the accuracy of trajectory predictions we adopt a suite of two metrics, which are used in many previous trajectory prediction work~\cite{ettinger2021large}.  We use $\hat{p} = \left\{\hat{p_t}\right\}$ to denote the ground truth point, where t is the time we desire to predict the position of the ego flight and we use $p = \left\{p_t\right\}$ to denote the predicted point.
\begin{itemize}
    \item Average Displacement Error (ADE): The ADE is the average L2 norm distance between the points in the ground truth trajectory and the points in the predicted trajectory:
$\frac{1}{k}  \sum_t\left\|\hat{p}_{t}-p_{t}\right\|_2$ where $t\in{t_{n+1}, t_{n+2}...t_{n+k}}$, k is the desired time step we want to predict

    \item Final Displacement Error (FDE): The FDE is the L2 norm between the last point in the ground truth trajectory and the last point in the predicted trajectory:
    $\left\|\hat{p}_{t_{n+k}}-p_{t_{n+k}}\right\|_2 $
    
\end{itemize}

\section{Results}
\subsection{Data Description}
The TrajAir dataset\cite{patrikar2022predicting} is a pioneering collection that captures the multiple flight paths around a typical non-towered airport(Pittsburgh-Butler Regional Airport (ICAO:KBTP)), complemented by the accompanying weather conditions. The FAA has set forth specific guidelines that aircraft must adhere to when entering or departing from non-towered airspace, ensuring the efficiency and safety of all involved parties. Notably, both runways of KBTP follow Left Traffic patterns. These traffic patterns take a rectangular form, characterized by left-turning maneuvers in relation to landing or takeoff directions. The dataset contains multiple scenes which are formed by multiple frames. Each frame of the dataset contains frame number, aircraft ID, and aircraft position at x, y, and z dimensions. and wind speed at the x and y dimensions. 
% \subsection{ADE and FDE comparison}

\begin{figure*}[htbp]
    \centering
    \includegraphics[width=\linewidth]{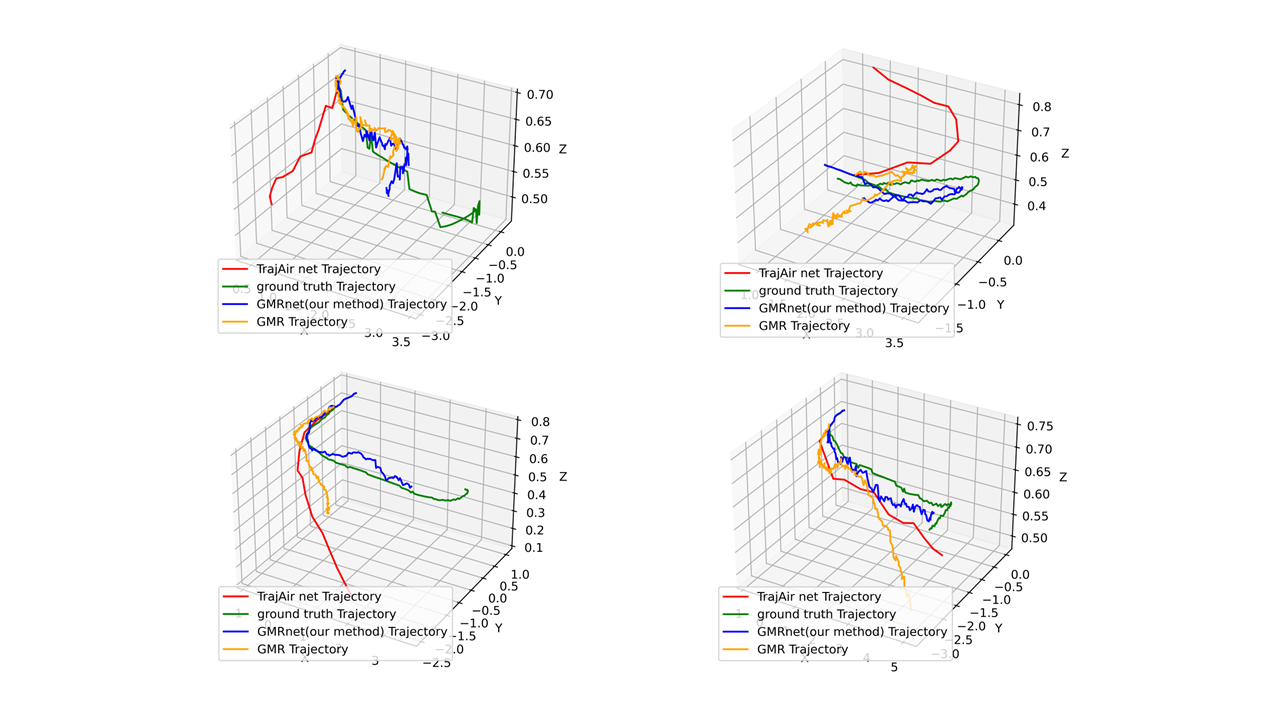}
    \caption{Examples of prediction comparison for each method.}
    \label{fig:ps}
\end{figure*}

\begin{figure*}[htbp]
    \centering
    \centering
    \includegraphics[width=\linewidth]{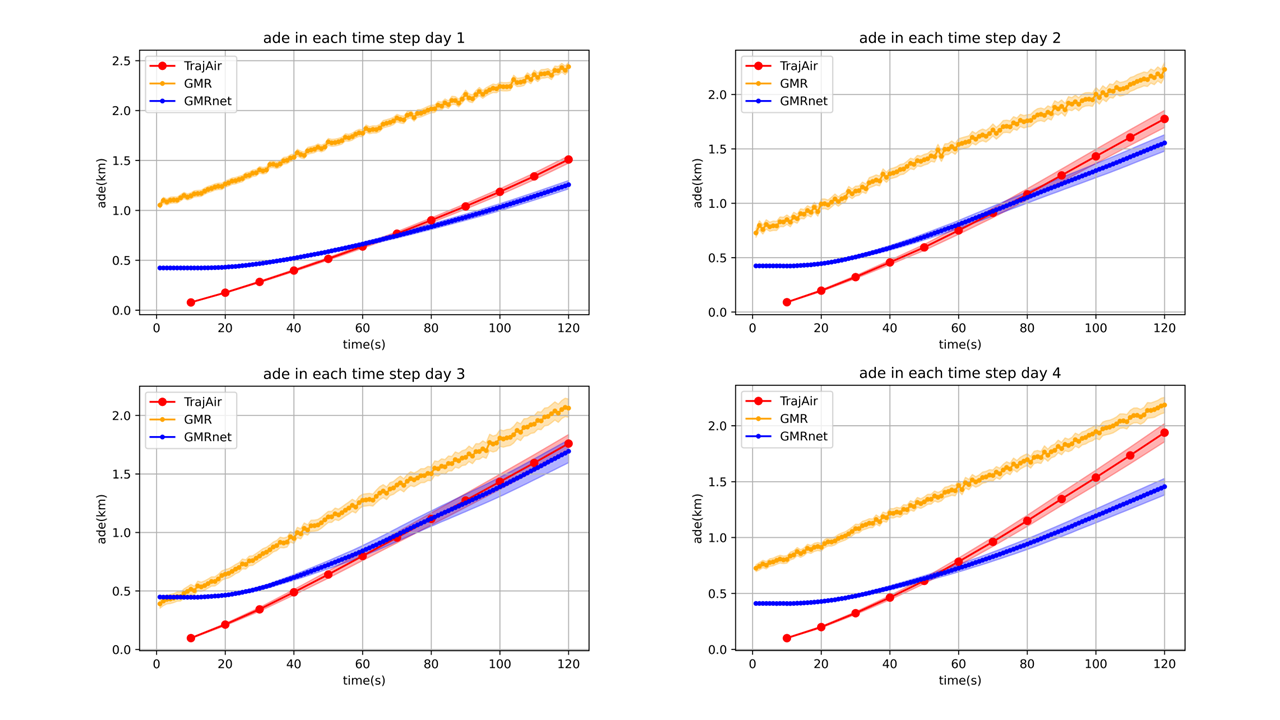}
    \caption{ADE results for each method.}
    \label{fig:result}
\end{figure*}
The evaluations are conducted using a 28-day segment from the TrajAir dataset. This subset encompasses four groups, each containing 7 sequential days from different months. We present the results from our model alongside comparisons with other methods across these four segments.  Day 1 includes 7143 predictable agents, Day 2 includes 2932 predictable agents, Day 3 includes 1936 predictable agents, and Day 4 includes 3225 predictable agents. With the aim of creating long-horizon predictions, we use the length of past trajectory n = 11 sec, guide time interval $\Delta$t = 10 sec, and predicting length k = 120 sec. Therefore, the prediction model observes 11 seconds of the past trajectory and predicts 120 seconds of the future trajectory with 12 guided points. The learning rate of the TrajAirNet is 1e-3, and the optimizer is Adam optimizer. The number of individual Gaussian is 150. We report the best ADE and FDE values of 5 samples in each test case for all methods.

\subsection{ADE and FDE comparison}
Figure~\ref{fig:ps} shows samples of predicting generated by our method and baseline methods compared to ground truth.
The ADE for each baseline method and our method on each dataset are in shown table~\ref{tab:ADE}. The results show that our method yields a considerably lower ADE compared to GMR. The ADE of our method is only 41\% on day 1, 58\% on day 2, 74\% on day 3, and 55\% on day 4 in relation to GMR's ADE while the variance is also reduced. The ADE of our method is quite close to the TrajAirNet, even though TrajAirNet only generates a 12-point prediction while our method generates a 120-point prediction.

The FDE for each baseline method and our method on each dataset in shown as table~\ref{tab:FDE}.
 our method also demonstrates a markedly lower FDE relative to GMR. The ADE of our method is merely 52\% on day 1, 69\% on day 2, 82\% on day 3, and 67\% on day 4 juxtaposed with GMR's FDE. Moreover, our method marginally surpasses TrajAirNet in terms of ADE.

The ade in each time step is shown as Figure~\ref{fig:result}.
These outcomes indicate our method offers a more accurate prediction than GMR. Furthermore, the ADE of our method is a little higher than the TrajAir at the first couple of time steps, but ultimately overpasses the TrajAir. It means our method has the same level of accuracy as TrajAir while tenfold the predicting temporal resolution.  
\begin{table*}[]
    \centering
    \caption{ADE Result}
\begin{tabular}{|c|c|c|c|c|}
\hline
\multicolumn{5}{|c|}{ADE(km)} \\
\hline Method & day1& day2 & day3 &day4\\
\hline GMR-net  (Ours) & 0.72$\pm$0.68 & 0.87$\pm$0.8 & 0.92$\pm$0.8 & 0.8$\pm$0.78 \\
\hline GMR   &  1.77$\pm$1.21 & 1.5$\pm$1.11 & 1.24$\pm$0.93 & 1.45$\pm$1.07\\
\hline TrajAir10  & 0.74$\pm$0.64 & 0.87$\pm$0.84 & 0.89$\pm$0.7 & 0.93$\pm$0.87\\
\hline
TrajAir1  & 5.57$\pm$3.92 & 7.33$\pm$8.14 & 12.61$\pm$11.73 & 4.68$\pm$3.79\\
\hline
\end{tabular}
    \label{tab:ADE}
\end{table*}
\begin{table*}[]
    \centering
    \caption{FDE Result}
\begin{tabular}{|c|c|c|c|c|}
\hline
\multicolumn{5}{|c|}{FDE(km)} \\
\hline Method & day1& day2 & day3 &day4\\
\hline GMR-net  (Ours) & 1.26$\pm$1.38 & 1.55$\pm$1.59 & 1.69$\pm$1.67 & 1.45$\pm$1.63 \\
\hline GMR   &  2.44$\pm$1.47 & 
2.23$\pm$1.4& 2.06$\pm$1.34 & 2.18$\pm$1.52\\
\hline TrajAir10  & 1.51$\pm$1.21 & 1.78$\pm$1.66 & 1.76$\pm$1.32 & 1.94$\pm$1.8 \\
\hline
TrajAir1  & 12.33$\pm$12.33 & 15.55$\pm$16.48 & 26.42$\pm$22.84 & 10.6$\pm$7.73\\
\hline
\end{tabular}
    \label{tab:FDE}
\end{table*}

% \begin{figure*}
%     \centering
%     \includegraphics[width=\textwidth]{figure/ade1.png}
%     \caption{ade day1}
%     \label{fig:day1ade}
% \end{figure*}
% \begin{figure*}
%     \centering
%     \includegraphics[width=\textwidth]{figure/ade2.png}
%     \caption{ade day2}
%     \label{fig:day2ade}
% \end{figure*}
% \begin{figure*}
%     \centering
%     \includegraphics[width=\textwidth]{figure/ade3.png}
%     \caption{ade day3}
%     \label{fig:day3ade}
% \end{figure*}
% \begin{figure*}
%     \centering
%     \includegraphics[width=\textwidth]{figure/ade4.png}
%     \caption{ade day4}
%     \label{fig:day4ade}
% \end{figure*}
 \subsection{Time  comparison}
The result for computing time is shown as table~\ref{tab:time}. The result shows that even though our method is apparently slower than the baseline methods, the computing time is still smaller in magnitude. The computing time for a one-second prediction only takes about 0.01 seconds.

\begin{table*}[]
    \centering
    \caption{Computing Time Result}
\begin{tabular}{|c|c|c|c|c|}
\hline
\multicolumn{5}{|c|}{average generating time per time step(second)} \\
\hline Method & day1& day2 & day3 &day4\\
\hline GMR-net  (Ours) & 0.010 & 0.011 & 0.016 & 0.013 \\
\hline GMR   &  0.005 & 0.007 & 0.010 & 0.008\\
\hline TrajAirNet  & 0.001 & 0.001 & 0.002 & 0.001 \\
\hline
\end{tabular}
    \label{tab:time}
\end{table*}
\section{Conclusion and Future work}
In conclusion, this paper proposed a learning framework consisting of four main modules. Each individual module collaborates with each other to achieve the best result because the strength of a module can compensate for the weaknesses of the others. The feature extraction module takes raw data directly, transforming it into more informative features. The feature fusion module can combine extracted features from different sources and formats together. The Guide generation module can utilize these fused features, which store all the provided information, and generate a low-dimensional guide for the mixture model module, who finally ultimately the final trajectory prediction. Each of the modules can be trained to enhance their performance in their respective tasks.
We then test the framework by filling the modules with some previous methods. The result shows that our framework outperforms the current state-of-the-art methods on the tested dataset.
In the future, the current framework is worthy to be exploited. For example, the framework may perform better with a stronger context encoder filled in the feature extraction module. We also would like this framework to be tested on different trajectory prediction datasets. In this work, we only test the proposed framework on one dataset because there is very few open-source flight trajectory dataset.

\newpage
\bibliographystyle{aiaa}
\bibliography{references}

\end{document}